\documentclass[conference]{IEEEtran}
\IEEEoverridecommandlockouts

\usepackage{cite}
\usepackage{amsmath,amssymb,amsfonts}
\usepackage{algorithmic}
\usepackage{graphicx}
\usepackage{textcomp}
\usepackage{xcolor}
\usepackage{booktabs}
\usepackage{url}
\usepackage{multirow}
\def\BibTeX{{\rm B\kern-.05em{\sc i\kern-.025em b}\kern-.08em
    T\kern-.1667em\lower.7ex\hbox{E}\kern-.125emX}}
\begin{document}

\title{\textsc{Text2Graph}: Combining Lightweight LLMs and GNNs for Efficient Text Classification in Label-Scarce Scenarios}

\author{\IEEEauthorblockN{João Lucas Luz Lima Sarcinelli}
\IEEEauthorblockA{\textit{ICMC-USP} \\
São Carlos, Brazil \\
joao.luz@usp.br} \and

\IEEEauthorblockN{Ricardo Marcondes Marcacini}
\IEEEauthorblockA{\textit{ICMC-USP} \\
São Carlos, Brazil \\
ricardo.marcacini@usp.br}
}

\maketitle

\begin{abstract}
Large Language Models (LLMs) have become effective zero-shot classifiers, but their high computational requirements and environmental costs limit their practicality for large-scale annotation in high-performance computing (HPC) environments. To support more sustainable workflows, we present \textsc{Text2Graph}, an open-source Python package that provides a modular implementation of existing text-to-graph classification approaches. The framework enables users to combine LLM-based partial annotation with Graph Neural Network (GNN) label propagation in a flexible manner, making it straightforward to swap components such as feature extractors, edge construction methods, and sampling strategies. We benchmark \textsc{Text2Graph} on a zero-shot setting using five datasets spanning topic classification and sentiment analysis tasks, comparing multiple variants against other zero-shot approaches for text classification. In addition to reporting performance, we provide detailed estimates of energy consumption and carbon emissions, showing that graph-based propagation achieves competitive results at a fraction of the energy and environmental cost.
\end{abstract}


\section{Introduction}

Large Language Models (LLMs) have shown to be powerful tools for a variety of natural language processing tasks \cite{zhang2025pushing, enis2024llmnmtadvancinglowresource, wang2024llm}. Among their capabilities, their use as zero-shot text classifiers are especially highlighted. Without requiring task-specific fine-tuning, LLMs can leverage their pre-training knowledge on previously unseen data in order to provide annotations with minimal supervision \cite{zhao2025surveylargelanguagemodels}. This has made them useful for scenarios where labeled data is scarce or expensive to obtain.

However, despite their flexibility, LLMs demand substantial computational resources. Running inference on modern LLMs often requires GPUs or TPUs with high memory capacity while consuming a considerable amount of resources, and the cost grows with both model size and dataset scale. The alternative cloud-based options pose additional financial costs through paid API access, as well as a more general environmental impact in the form of large, resource intensive data-centers necessary to host and operate the underlying LLMs \cite{jegham2025hungryaibenchmarkingenergy}. When applied to large databases, the expense of using LLMs for annotation may become prohibitive. Moreover, modern LLMs require vast computational resources, both for training and inference, often exceeding the memory and energy budgets of high-performance computing (HPC) environments. This computational bottleneck raises important questions about the sustainability and feasibility of their use in large-scale labeling tasks.

Although current LLM approaches continue to evolve toward larger scales and increasing numbers of parameters, true efficiency in machine learning arises from how well data representations capture relevant structure and generalize beyond training. In this context, an alternative lies in graph-based approaches, which offer an efficient way to represent and propagate global knowledge across interconnected data points \cite{hoang2023graphrepresentationlearning}. Graphs naturally encode relational information and allow global integration of knowledge by connecting objects through edges that capture similarity or other forms of dependence. In this case, Graph Neural Networks (GNNs) present themselves as a way to integrate deep learning to these structures, providing mechanisms to infer missing labels and generalize annotations across the dataset in a resource-efficient manner \cite{wang2021combining}. Although not all databases are found in a structured setting, that is, in graph or graph-friendly form, a graph structure may still be constructed from the documents textual data \cite{haiderrizvi2025textclassificationusing}.

In this work, we propose \textsc{Text2Graph}, a highly modular and customizable pipeline that combines LLMs and GNNs for the unstructured text classification task. We first convert the individual text documents into a graph structure. Then, we use LLMs to annotate a sample of the large database. A GNN is then used to propagate the generated labels to the remaining unlabeled nodes. By doing so, we reduce the reliance on expensive LLM inference while still making use of their generalization capabilities to guide the learning process. Code for this pipeline is made publicly available as a Python package.

Finally, we evaluate this approach by comparing the resource consumption of pure LLM annotation against the proposed pipeline and variants. Our results highlight how integrating LLM-based partial annotation with GNN-based propagation can reduce use of resources, such as consumed energy and carbon emissions, while maintaining competitive labeling performance. This study contributes to the growing interest in scalable and efficient methods for large-scale data annotation, focusing on the role of graphs as an alternative to the use of large, resource consuming models. The main contributions of this paper are:

\begin{itemize}
    \item A unified pipeline for converting unstructured text classification problems into graph text classification problems. An implementation of the proposed pipeline is also made publicly available as a Python library;
    \item An in-depth performance comparison of different variations of the proposed pipeline, including methods found in the literature, for the zero-shot, label-free graph text classification problem. Tests are conducted using five different datasets, covering three distinct tasks;
    \item Resource utilization comparisons for different zero-shot text classification approaches. We analyze overall energy consumption and carbon emissions for different techniques.
\end{itemize}

\section{Related Works}



\subsection{Graph-based Text Classification}

Text classification through graphs may be divided into two main categories: classification on Text Attributed Graphs (TAGs), which are graph structures that have relevant node-level textual attributes \cite{jin2024largelanguagemodels}, such as social networks or citation networks; and classification on unstructured textual data, which is textual data that is not previously available in graph form, such as news articles and product reviews. In this work we'll be focusing on the unstructured textual data classification problem. 

In order to operate through graphs, a graph structure must be inferred from the available unstructured textual data \cite{wang2024graphneuralnetworks}. Previous works explore constructing heterogeneous document-word graphs \cite{yao2019graphconvolutionalnetworks}, with nodes representing both documents and individual words. Edges may be created using different techniques, such as TF-IDF and point-wise multiple information (PMI) \cite{wang2024graphneuralnetworks}. Homogeneous graphs are also explored, in which only documents are represented as nodes. \cite{benamira2019semisupervisedlearninggraph} constructs a graph from unstructured news articles using a $k$-Nearest Neighbors (kNN) approach. Each document is converted into a node, with edges connecting to the $k$ nearest nodes based on the euclidean distance of extracted document embeddings. 

\subsection{LLMs for Text Classification with Graphs}

LLMs may be used for textual graph problems in two main forms \cite{chen2024exploringpotentiallarge}: as \textbf{predictors}, in which they are the system's final classifier and must use the graph structure to this end; and as \textbf{enhancers}, when they must be used to enrich the graph's textual data for later use in classification, commonly done through GNNs. While many works focus on incorporating the graph structure inside LLM prompts for better classification performance \cite{hu2024letsaskgnn, chen2024exploringpotentiallarge, ye2024language}, we'll focus on their use as enhancers.

In \cite{chen2023labelfreenodeclassification}, the authors perform label-free node classification on TAGs by sampling nodes based on clustering of textual embedding and generates labels using a series of prompts to GPT-3.5. The responses are then used to annotate the sampled nodes along with confidence scores, which are then propagated to the unlabeled nodes using a Graph Convolutional Network (GCN). \cite{he2023harnessingexplanationsllmtolm} classifies TAG nodes using the LLM while generating explanations. They then train interpreter language models on the original texts and explanations to perform the classification task. They also prompt the model to generate a list of top-k ranked predictions, which are then used to train three GNNs. All outputs are combined using ensembling to generate the final classification.


\section{\textsc{Text2Graph} Pipeline}

\begin{figure*}[htpb]
    \centering
    \includegraphics[width=0.9\linewidth]{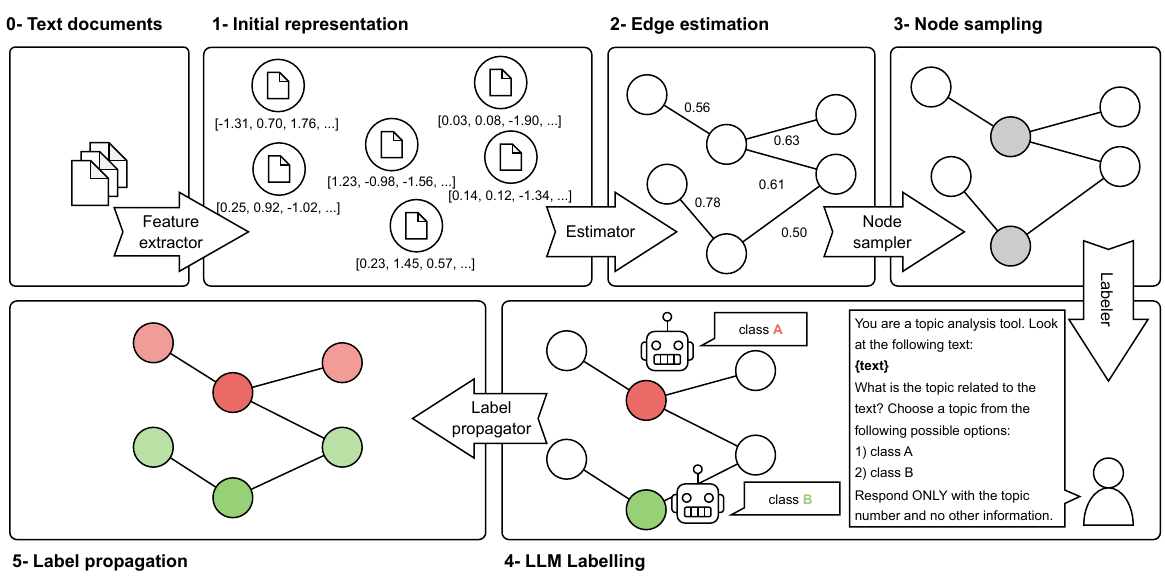}
    \caption{The \textsc{Text2Graph} pipeline. Textual documents have initial representations extracted and edges are created accordingly. Then, some nodes are selected to be labeled by the LLM. The labels are then propagated to the unlabeled nodes using a GNN.}
    \label{fig:pipeline}
\end{figure*}

In this section we describe the \textsc{Text2Graph} pipeline\footnote{Available at \url{https://github.com/Joao-Luz/Text2Graph}}, which converts the text classification task into a node classification task by building a graph from the raw text documents. Each document is represented as a single node in our graph and classification is done in a zero-shot setting, that is, we assume no ground truth labels are available. Our approach is similar to the one shown in \cite{benamira2019semisupervisedlearninggraph}. An overview of the pipeline is shown in Figure \ref{fig:pipeline}.

The pipeline has five main steps, which are executed by five distinct modules: \textbf{initial representation extraction}, executed by the feature extractor module; \textbf{edge estimation}, executed by the estimator module; \textbf{node sampling}, executed via the node sampler module; \textbf{LLM labeling}, done by the labeler module; and \textbf{label propagation}, via the label propagator module. We'll now describe each step functionality in detail:
\vspace{1em}

\noindent\textbf{Initial representation}\hspace{1em} The first step of the pipeline is to extract an initial embedding representation for each node/document. This is done using the \textbf{feature extractor} module which, in our pipeline, the default option is the sentence embedding extractor model \texttt{all-MiniLM-L6-v2}\footnote{\url{https://huggingface.co/sentence-transformers/all-MiniLM-L6-v2}}, although users may replace it with any other compatible embedding model. The representation is later used for both constructing edges and as node representations for a GNN model.

\noindent\textbf{Edge estimation}\hspace{1em} This step connects the document nodes to each other using the \textbf{estimator} module. In our configuration, this is done by computing the cosine distance matrix $D$ between each pair of embedding representations in the graph. Next, the Minimum Spanning Tree (MST) of the graph is computed, using $D$ as an adjacency matrix, and the graph edges are set to that of the MST, along with the distances representing the edge weights. By building the graph from the MST, we aim to make it connected, which may have an impact on the label propagation step.


\noindent\textbf{Node sampling}\hspace{1em} As we aim to propagate LLM generated labels to unlabeled graph nodes, we must first select the nodes that will be labeled by the LLM. The node sampling step selects these nodes using the \textbf{node sampler} module. In our configuration, we randomly select $10\%$ of the available nodes.

\noindent\textbf{LLM labeling}\hspace{1em} LLM labeling is done by the \textbf{labeler} module, which labels the nodes selected in the previous step of the pipeline. We use the LLaMA3.1-8B\footnote{\url{https://huggingface.co/meta-llama/Llama-3.1-8B-Instruct}} \cite{grattafiori2024thellama3} as the generative model, keeping default decoding setting for the model and temperature at $t=0.0$. The prompt used, shown in Figure \ref{fig:pipeline}, is a multiple-choice, zero-shot prompt which contains the general task description, the node's textual content and the class options available for the task.


\noindent\textbf{Label propagation}\hspace{1em} After labels are obtained in the previous step by the labeler module, they are propagated to the remaining nodes in the label propagation step via the \textbf{label propagator} module. For our pipeline, we make use of a simple two-layer GCN with 16 hidden channels. The model is trained over 500 epochs with a learning rate $lr=10^{-3}$ and, after training, we use this model to infer the labels for the remaining unlabeled nodes.

\vspace{1em}
By using a highly modular framework, its possible to integrate different components to the pipeline and easily assess their impact on the final classification task. For example, the estimator module may be changed so edges are constructed using a different method, such as the kNN approach shown in \cite{benamira2019semisupervisedlearninggraph}, or the node sampler may be exchanged from a random selector to one that selects nodes with highest degrees.

\section{Experimental Setup}

\subsection{Datasets}

We select five common datasets used for unstructured text classification using graphs, based on the findings of recent surveys \cite{haiderrizvi2025textclassificationusing, wang2024graphneuralnetworks}. The datasets are:

\begin{itemize}
    \item \textbf{Ohsumed} \cite{joachims1998textcategorization}: The original Ohsumed dataset is a multi-label classification dataset on academic papers. The task is to classify to which topics a given document is associated, from a list of 23 possible topics. A commonly used version of this dataset is the subset of documents associated with only one label \cite{yao2019graphconvolutionalnetworks}, making it a single-label task. This is the version we'll be using;
    \item \textbf{Reuters 8/52}: These are two subsets of the main Reuters 21587 financial news dataset\footnote{\url{http://www.daviddlewis.com/resources/testcollections/reuters21578}}, which compiles news articles and classifies them into 8 and 52 possible topics, respectively;
    \item \textbf{AG News} \cite{zhang2015characterlevel}: Another dataset for topic classification in news articles. This is a general domain dataset, with articles classified into four categories;
    \item \textbf{Internet Movie Database (IMDB)} \cite{maas2011learningword}: This is a sentiment analysis dataset of movie reviews. It is a binary classification dataset composed of relatively small textual documents.
\end{itemize}

As we aim to evaluate zero-shot text classification using graphs, we perform our tests on the respective test sets of each dataset and make no use of train data. A summary of each dataset is shown in Table \ref{tab:datasets}.

\begin{table}[htpb]
    \centering
    \caption{Summary of the datasets.}
    \label{tab:datasets}
    \begin{tabular}{lllcc}
        \toprule
        \textbf{Dataset} & \textbf{Task} & \textbf{Domain} & \textbf{\# classes} & \textbf{\# docs} \\
        \midrule
        Ohsumed & \multirow{4}{*}{Topic class.} & Medical bib. & 23 & 4040 \\
        Reuters 8 &  & Financial news & 8 & 2189 \\
        Reuters 52 &  & Financial news & 52 & 2568 \\
        AG News &  & General news & 4 & 7600 \\
        IMDB & Sentiment & Movie reviews & 2 & 25,000 \\
        \bottomrule
    \end{tabular}
\end{table}

\subsection{Methods}

We conduct experiments using variations of the \textsc{Text2Graph} pipeline. This is done by modifying the modules already in use. We investigate the following variants:

\begin{itemize}
    \item Using a kNN approach to generating edges between the graph nodes, similar to what is done in \cite{benamira2019semisupervisedlearninggraph}. We use $k=5$. Its important to note that this approach doesn't guarantee that the final graph is connected;
    \item Changing the random sampling to a sampling based on the nodes' degree. The idea is that these nodes will be able to pass information to more neighbors during propagation, thus making them more relevant to be initially labeled.
    \item Employ a larger version of the base LLM as the annotator. Namely, we use LLaMA3.1-70B\footnote{\url{https://huggingface.co/meta-llama/Llama-3.1-70B-Instruct}} as the larger model.
    \item Making use of ground truth labels in place of LLM labeling, to simulate the use of a human annotator.
\end{itemize}

Furthermore, we also test two strategies that differ from the original pipeline's structure:

\begin{itemize}
    \item The use of the LLM as a full labeler, using the same prompting strategy as in the labeling step of the pipeline. We aim to compare the cost-effectiveness in the extreme scenario where only the LLM is used for classification;
    \item Training a DistilBERT \cite{sanh2020distilbertdistilledversionbert} model\footnote{\url{https://huggingface.co/docs/transformers/en/model_doc/distilbert}} on the sampled LLM annotations and using its predictions, using 10 epochs for training, learning rate $lr=5\times10^{-5}$. This is done to compare the use of the graph modeling to a traditional text classification pipeline in the zero-shot setting. 
\end{itemize}




\subsection{Hardware Configurations}

Experiments were run using a NVIDIA GeForce RTX 4090 graphics card, which has a capacity of 24GB of VRAM, coupled with a 13th Gen Intel(R) Core(TM) i5-13600KF and 64GB of RAM. Experiments were executed in Brazil's São Paulo state, which is relevant for the estimation of CO2 emissions during the experiments\footnote{\url{https://codecarbon.io/#howitwork}}. LLMs were ran using the vLLM Python package\footnote{\url{https://docs.vllm.ai}}. LLaMA3.1-8B was fully ran on the GPU memory, while LLaMA3.1-70B had to be quantized using BitsAndBytes 4bit quantization\footnote{\url{https://huggingface.co/unsloth/Meta-Llama-3.1-70B-Instruct-bnb-4bit}}, and ran partially on the regular RAM.

\section{Results}

\subsection{Overall Performance}

\begin{figure}
    \centering
    \includegraphics[width=1.0\linewidth]{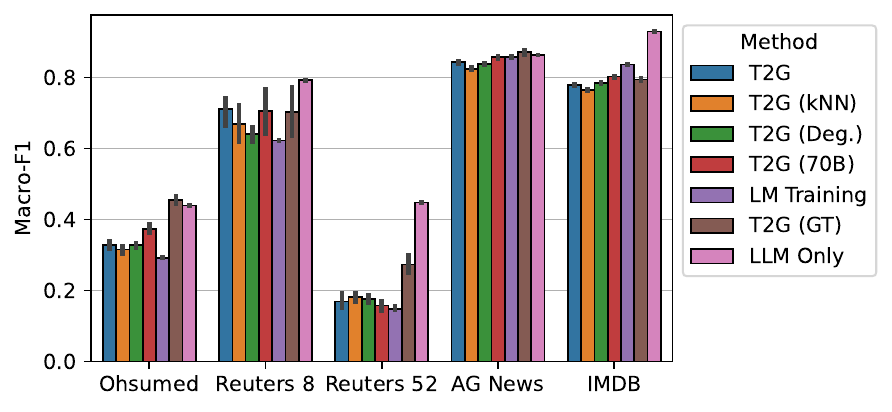}
    \caption{Overall performance of methods for the different datasets.}
    \label{fig:overall}
\end{figure}

We first evaluate the overall performance of the methods for the classification tasks. For the analysis, we'll consider the F1-Macro score. Results are illustrated in Figure \ref{fig:overall}.

Full LLM labeling (LLM Only) achieves higher overall performance across all datasets, achieving equivalent scores or even surpassing the original \textsc{Text2Graph} pipeline that uses ground truth labels (T2G GT). Nevertheless, the pipeline remains competitive in most cases, provided the underlying LLM performs sufficiently well in the labeling step. Using a stronger LLM (T2G (70B)) did not appear to make a noticeable difference in the pipeline's performance; in fact, both Reuters datasets show a slight decrease. This indicates that the larger model's performance is not sufficiently better to justify its use.

Using kNN edge construction (T2G (kNN)) was not significantly more beneficial than using the MST. This may be because the kNN approach does not guarantee a connected graph, which could hinder the GCN's ability to propagate labels in some cases. Sampling by the nodes' degrees shows a similar trend, with no expressive gains and decrease in performance in some cases compared to random sampling.

In settings where overall performance is low, the gap between the original pipeline and the use of ground truth labels widens. In higher-performing settings, however, this gap narrows. This suggests that while graph modeling is beneficial, the quality of the underlying annotations is also critical. When LLMs perform well as annotators, there is no significant performance loss compared to using a human annotator.

For topic classification datasets with fewer classes (AG News and Reuters 8), overall performance is higher and the performance gap between LLM labeling and graph-based approaches is smaller. For datasets with more classes, such as Ohsumed (23) and Reuters 52 (52), this gap widens. One explanation is that graph-based methods only annotate a small data sample. If examples from under-represented classes are not selected, or are incorrectly classified, those labels will be absent from the graph and cannot be propagated, severely impacting performance for those classes. As shown in Table \ref{tab:missing}, class representation is a significant challenge: for Ohsumed, some classes are frequently missing from the final graph, and for Reuters 52, 18 classes are always missing while 8 are missing 80\% of the time. Changing the sampling strategy from random to degree based reduces the amount of missing classes for Reuters 8 and Ohsumed, but it increases the number of classes missing 100\% of the time for Reuters 52.

\begin{table}[htpb]
    \centering
    \caption{Number of runs class labels are missing from the graphs varying node sampling.}
    \label{tab:missing}
    \begin{tabular}{lccc}
        \toprule
        \multirow{2}{*}{\textbf{Dataset}} & \multirow{2}{*}{\textbf{\# times missing}} & \multicolumn{2}{c}{\textbf{\# classes}} \\
        \cmidrule{3-4}
         & & Random & Degree \\
        \midrule
        Reuters 8 & 1/5 & 1 & - \\
        \midrule
        \multirow{5}{*}{Reuters 52} & 5/5 & 18 & 29\\
         & 4/5 & 8 & - \\
         & 3/5 & 2 & - \\
         & 2/5 & 6 & - \\
         & 1/5 & 8 & - \\
        \midrule
        \multirow{3}{*}{Ohsumed} & 5/5 & 1 & 1\\
         & 2/5 & 1 & -\\
         & 1/5 & 2 & -\\
        \bottomrule
    \end{tabular}
\end{table}

Lastly, in most cases, using graphs to capture global relations between data points are as effective or even better than training a specialized language model on the LLM's annotations (LM Training). For the sentiment analysis task, however, LM training comes in as a strong second after pure LLM labeling when compared to the graph-based methods, even surpassing the use of ground-truth as labels. This may be explained by the nature of the task, in which the order of words within sentences is of greater importance of how individual reviews relate to each other globally. This shows that some tasks have more to gain from the graph modeling than others.

\subsection{Resource Use}

\begin{figure}[htpb]
    \centering
    \includegraphics[width=1.0\linewidth]{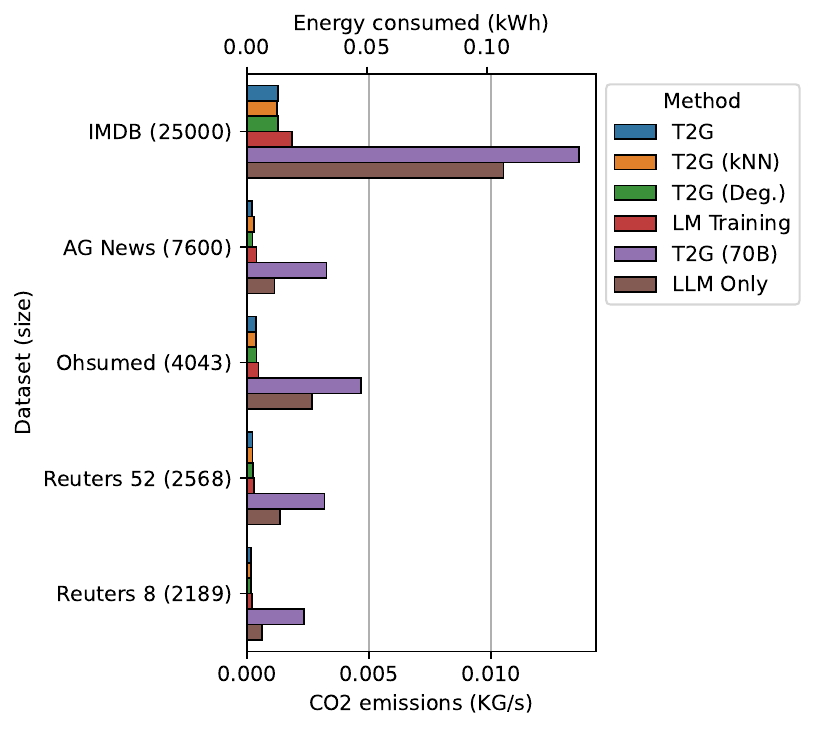}
    \caption{Average equivalent CO2 emissions and energy consumed across runs.}
    \label{fig:energy}
\end{figure}

Using the \texttt{codecarbon} Python library, we're able to estimate the energy consumption and equivalent CO2 emissions when using the different methods for the classification task. Estimates for the different methods are shown in Figure \ref{fig:energy}.

The consumed energy has direct relation with dataset size and average document length. The energy consumption and CO2 emissions of the graph based methods are considerably lower than that of the full LLM labeling, even in cases where the performance gap between the two methods is not large. For larger databases, as illustrated by IMDB, the use of a distillation method is strongly justified, as they output similar performances at a fraction of the power and environmental cost. Furthermore, the use of the larger model is further discouraged in light of its operational cost: using it to label 10\% of the database consumes more energy than the full annotation done by a smaller model. Although, it is important to note that, as experiments were conducted using limited hardware, the larger model was required to run partially on the regular RAM, which may impact the runtime and, thus, energy and CO2 estimates. Nonetheless, as this setup may also be found in real world applications, the comparison may still be valid.

\begin{figure}[htpb]
    \centering
    \includegraphics[width=1.0\linewidth]{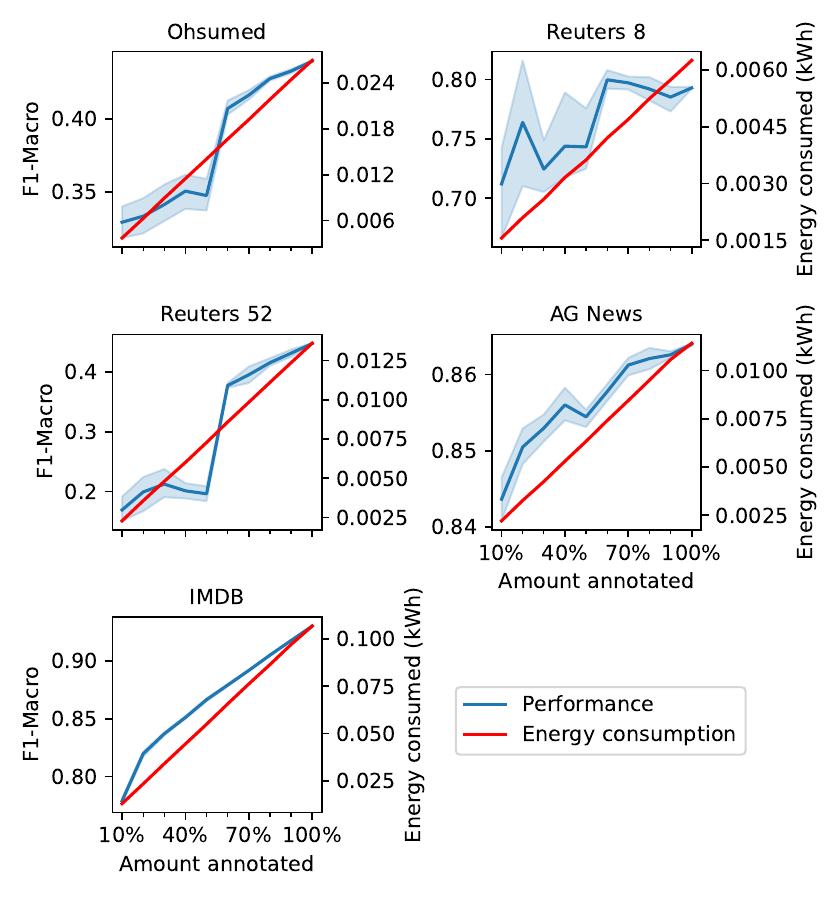}
    \caption{Performance comparison to energy consumption when increasing the amount of sampled nodes for annotation.}
    \label{fig:ablation}
\end{figure}

By increasing the amount of documents initially annotated by the LLM, we can see how energy consumption relates to the system's overall performance when the LLM is used to a greater extent. This comparison is shown in Figure \ref{fig:ablation}, which shows how the energy grows linearly with the amount of labeled documents. In most cases, however, the performance doesn't follow the same linear behavior, showing performance decreases and diminishing returns, depending on the dataset. Furthermore, even though performance tends to increase with the amount of documents processed by the LLM, larger samples may not always be justified. For AG News, for example, performance grows by roughly 0.02, while energy consumption more than quadruples. These results call attention to the question of ``what is an acceptable trade-off between energy consumption and model performance?''

\begin{figure}[htpb]
    \centering
    \includegraphics[width=1.0\linewidth]{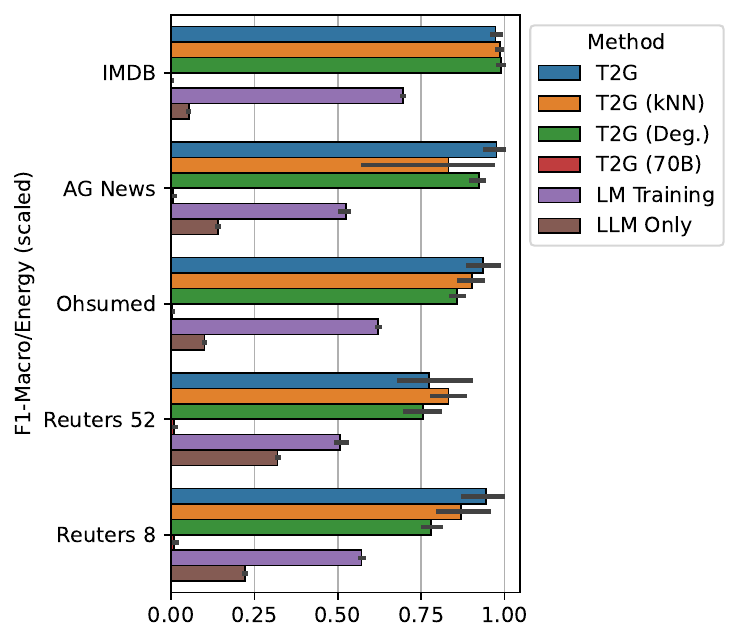}
    \caption{Average min max-scaled ratio of performance over consumed energy.}
    \label{fig:performance_energy_ratio}
\end{figure}

The ratio of the final F1 score over the respective consumed energy is shown in Figure \ref{fig:performance_energy_ratio}. Here, the values were scaled using Min Max Scaling to make the ratios more comparable across datasets. We can see that, by this heuristic, graph based methods are the most cost-efficient methods for the task, with the original \textsc{Text2Graph} pipeline often achieving the highest mark when compared to its variations, although they all tend to perform similarly in this comparison. LM training comes in second, scoring lower due to its computational overhead, seen in Figure \ref{fig:energy}. Full LLM annotations are second to last due to the high power consumption, even though performance tends to be higher, while the use of the larger LLM neither consumes less energy nor performs better than others, cementing it as the least cost-efficient method.

\section{Conclusion}

In this work we evaluate the zero-shot text classification task using graph modeling and LLMs as annotators. The unified pipeline \textsc{Text2Graph}, which is highly modular and customization, is made available publicly as a Python package. Experiments were conducted using the pipeline and variations across five datasets spanning three text classification tasks. We found that using graph modeling may be beneficial for final performance in some of the tasks, especially when the underlying LLM performs sufficiently well in the zero-shot classification task, although challenges regarding this approach are highlighted. Notably, when comparing the energy use and CO2 emissions of different methods, we see that the sole use of LLM labeling presents much higher costs when compared to methods that perform distillation, which can be unjustifiable for large scaled text classification applications.

The \textsc{Text2Graph} pipeline may be improved in order to tackle some of the challenges it faces. For example, different sampling strategies derived form active learning may be applied to mitigate the class representativeness problem faced in datasets like Reuters 52. Furthermore, alternative and more complex graph constructions may be investigated in order to provide a more meaningful global structure to the graph. Its important to highlight, though, that due to it being developed in a highly modular manner, these modifications may be easily integrated to the pipeline.

Another direction for future work is the exploration of committee-based labeling, where multiple lightweight LLMs are combined to generate pseudo-labels collaboratively. This ensemble strategy can reduce the bias of individual models and provide more reliable annotations for subsequent graph propagation. A further promising path is the adoption of a teacher–student framework, in which a larger LLM serves as a teacher to fine-tune smaller LLMs for specific domains. Although this approach may introduce higher training costs, especially on large datasets, the resulting student models could offer a favorable trade-off between accuracy and carbon emissions in repeated large-scale inference tasks.

\bibliographystyle{ieeetr}
\bibliography{references}

@misc{sanh2020distilbertdistilledversionbert,
    title={DistilBERT, a distilled version of BERT: smaller, faster, cheaper and lighter}, 
    author={Victor Sanh and Lysandre Debut and Julien Chaumond and Thomas Wolf},
    year={2020},
    eprint={1910.01108},
    archivePrefix={arXiv},
    primaryClass={cs.CL},
    url={https://arxiv.org/abs/1910.01108}, 
}

@inproceedings{benamira2019semisupervisedlearninggraph,
  title = {Semi-Supervised Learning and Graph Neural Networks for Fake News Detection},
  booktitle = {Proceedings of the 2019 {{IEEE}}/{{ACM International Conference}} on {{Advances}} in {{Social Networks Analysis}} and {{Mining}}},
  author = {Benamira, Adrien and Devillers, Benjamin and Lesot, Etienne and Ray, Ayush K. and Saadi, Manal and Malliaros, Fragkiskos D.},
  year = {2019},
  month = aug,
  pages = {568--569},
  publisher = {ACM},
  address = {Vancouver British Columbia Canada},
  doi = {10.1145/3341161.3342958},
  urldate = {2025-08-27},
  isbn = {978-1-4503-6868-1},
  langid = {english}
}

@misc{chen2023labelfreenodeclassification,
  title = {Label-Free {{Node Classification}} on {{Graphs}} with {{Large Language Models}} ({{LLMS}})},
  author = {Chen, Zhikai and Mao, Haitao and Wen, Hongzhi and Han, Haoyu and Jin, Wei and Zhang, Haiyang and Liu, Hui and Tang, Jiliang},
  year = {2023},
  publisher = {arXiv},
  doi = {10.48550/ARXIV.2310.04668},
  urldate = {2025-08-10},
  copyright = {arXiv.org perpetual, non-exclusive license}
}

@article{chen2024exploringpotentiallarge,
  title = {Exploring the {{Potential}} of {{Large Language Models}} ({{LLMs}})in {{Learning}} on {{Graphs}}},
  author = {Chen, Zhikai and Mao, Haitao and Li, Hang and Jin, Wei and Wen, Hongzhi and Wei, Xiaochi and Wang, Shuaiqiang and Yin, Dawei and Fan, Wenqi and Liu, Hui and Tang, Jiliang},
  year = {2024},
  month = mar,
  journal = {ACM SIGKDD Explorations Newsletter},
  volume = {25},
  number = {2},
  pages = {42--61},
  issn = {1931-0145, 1931-0153},
  doi = {10.1145/3655103.3655110},
  urldate = {2025-08-11},
  langid = {english}
}

@article{haiderrizvi2025textclassificationusing,
  title = {Text {{Classification Using Graph Convolutional Networks}}: {{A Comprehensive Survey}}},
  shorttitle = {Text {{Classification Using Graph Convolutional Networks}}},
  author = {Haider Rizvi, Syed Mustafa and Imran, Ramsha and Mahmood, Arif},
  year = {2025},
  month = aug,
  journal = {ACM Computing Surveys},
  volume = {57},
  number = {8},
  pages = {1--38},
  issn = {0360-0300, 1557-7341},
  doi = {10.1145/3714456},
  urldate = {2025-08-10},
  langid = {english}
}

@misc{he2023harnessingexplanationsllmtolm,
  title = {Harnessing {{Explanations}}: {{LLM-to-LM Interpreter}} for {{Enhanced Text-Attributed Graph Representation Learning}}},
  shorttitle = {Harnessing {{Explanations}}},
  author = {He, Xiaoxin and Bresson, Xavier and Laurent, Thomas and Perold, Adam and LeCun, Yann and Hooi, Bryan},
  year = {2023},
  publisher = {arXiv},
  doi = {10.48550/ARXIV.2305.19523},
  urldate = {2025-08-10},
  copyright = {arXiv.org perpetual, non-exclusive license}
}

@article{hoang2023graphrepresentationlearning,
  title = {Graph {{Representation Learning}} and {{Its Applications}}: {{A Survey}}},
  shorttitle = {Graph {{Representation Learning}} and {{Its Applications}}},
  author = {Hoang, Van Thuy and Jeon, Hyeon-Ju and You, Eun-Soon and Yoon, Yoewon and Jung, Sungyeop and Lee, O-Joun},
  year = {2023},
  month = apr,
  journal = {Sensors},
  volume = {23},
  number = {8},
  pages = {4168},
  issn = {1424-8220},
  doi = {10.3390/s23084168},
  urldate = {2025-08-26},
  copyright = {https://creativecommons.org/licenses/by/4.0/},
  langid = {english}
}

@article{jin2024largelanguagemodels,
  title = {Large {{Language Models}} on {{Graphs}}: {{A Comprehensive Survey}}},
  shorttitle = {Large {{Language Models}} on {{Graphs}}},
  author = {Jin, Bowen and Liu, Gang and Han, Chi and Jiang, Meng and Ji, Heng and Han, Jiawei},
  year = {2024},
  month = dec,
  journal = {IEEE Transactions on Knowledge and Data Engineering},
  volume = {36},
  number = {12},
  pages = {8622--8642},
  issn = {1041-4347, 1558-2191, 2326-3865},
  doi = {10.1109/TKDE.2024.3469578},
  urldate = {2025-08-10},
  copyright = {https://ieeexplore.ieee.org/Xplorehelp/downloads/license-information/IEEE.html}
}

@article{wang2024graphneuralnetworks,
  title = {Graph Neural Networks for Text Classification: A Survey},
  shorttitle = {Graph Neural Networks for Text Classification},
  author = {Wang, Kunze and Ding, Yihao and Han, Soyeon Caren},
  year = {2024},
  month = jul,
  journal = {Artificial Intelligence Review},
  volume = {57},
  number = {8},
  pages = {190},
  issn = {1573-7462},
  doi = {10.1007/s10462-024-10808-0},
  urldate = {2025-08-10},
  langid = {english}
}

@article{yao2019graphconvolutionalnetworks,
  title = {Graph {{Convolutional Networks}} for {{Text Classification}}},
  author = {Yao, Liang and Mao, Chengsheng and Luo, Yuan},
  year = {2019},
  month = jul,
  journal = {Proceedings of the AAAI Conference on Artificial Intelligence},
  volume = {33},
  number = {01},
  pages = {7370--7377},
  issn = {2374-3468, 2159-5399},
  doi = {10.1609/aaai.v33i01.33017370},
  urldate = {2025-08-28},
  copyright = {https://www.aaai.org}
}

@InProceedings{joachims1998textcategorization,
  author="Joachims, Thorsten",
  editor="N{\'e}dellec, Claire
  and Rouveirol, C{\'e}line",
  title="Text categorization with Support Vector Machines: Learning with many relevant features",
  booktitle="Machine Learning: ECML-98",
  year="1998",
  publisher="Springer Berlin Heidelberg",
  address="Berlin, Heidelberg",
  pages="137--142",
  isbn="978-3-540-69781-7"
}

@inproceedings{zhang2015characterlevel,
    author = {Zhang, Xiang and Zhao, Junbo and LeCun, Yann},
    title = {Character-level convolutional networks for text classification},
    year = {2015},
    publisher = {MIT Press},
    address = {Cambridge, MA, USA},
    booktitle = {Proceedings of the 29th International Conference on Neural Information Processing Systems - Volume 1},
    pages = {649–657},
    numpages = {9},
    location = {Montreal, Canada},
    series = {NIPS'15}
}

@inproceedings{maas2011learningword,
    title = "Learning Word Vectors for Sentiment Analysis",
    author = "Maas, Andrew L.  and
      Daly, Raymond E.  and
      Pham, Peter T.  and
      Huang, Dan  and
      Ng, Andrew Y.  and
      Potts, Christopher",
    editor = "Lin, Dekang  and
      Matsumoto, Yuji  and
      Mihalcea, Rada",
    booktitle = "Proceedings of the 49th Annual Meeting of the Association for Computational Linguistics: Human Language Technologies",
    month = jun,
    year = "2011",
    address = "Portland, Oregon, USA",
    publisher = "Association for Computational Linguistics",
    url = "https://aclanthology.org/P11-1015/",
    pages = "142--150"
}

@inproceedings{zhang2025pushing,
    author = {Zhang, Yazhou and Wang, Mengyao and Li, Qiuchi and Tiwari, Prayag and Qin, Jing},
    title = {Pushing The Limit of LLM Capacity for Text Classification},
    year = {2025},
    isbn = {9798400713316},
    publisher = {Association for Computing Machinery},
    address = {New York, NY, USA},
    url = {https://doi.org/10.1145/3701716.3715528},
    doi = {10.1145/3701716.3715528},
    booktitle = {Companion Proceedings of the ACM on Web Conference 2025},
    pages = {1524–1528},
    numpages = {5},
    location = {Sydney NSW, Australia},
    series = {WWW '25}
}

@misc{enis2024llmnmtadvancinglowresource,
    title={From LLM to NMT: Advancing Low-Resource Machine Translation with Claude}, 
    author={Maxim Enis and Mark Hopkins},
    year={2024},
    eprint={2404.13813},
    archivePrefix={arXiv},
    primaryClass={cs.CL},
    url={https://arxiv.org/abs/2404.13813}, 
}

@article{wang2024llm,
  title={Llm for sentiment analysis in e-commerce: A deep dive into customer feedback},
  author={Wang, Zeyu and Zhu, Yue and He, Shuyao and Yan, Hao and Zhu, Ziyi},
  journal={Applied Science and Engineering Journal for Advanced Research},
  volume={3},
  number={4},
  pages={8--13},
  year={2024}
}

@misc{zhao2025surveylargelanguagemodels,
      title={A Survey of Large Language Models}, 
      author={Wayne Xin Zhao and Kun Zhou and Junyi Li and Tianyi Tang and Xiaolei Wang and Yupeng Hou and Yingqian Min and Beichen Zhang and Junjie Zhang and Zican Dong and Yifan Du and Chen Yang and Yushuo Chen and Zhipeng Chen and Jinhao Jiang and Ruiyang Ren and Yifan Li and Xinyu Tang and Zikang Liu and Peiyu Liu and Jian-Yun Nie and Ji-Rong Wen},
      year={2025},
      eprint={2303.18223},
      archivePrefix={arXiv},
      primaryClass={cs.CL},
      url={https://arxiv.org/abs/2303.18223}, 
}

@misc{jegham2025hungryaibenchmarkingenergy,
      title={How Hungry is AI? Benchmarking Energy, Water, and Carbon Footprint of LLM Inference}, 
      author={Nidhal Jegham and Marwan Abdelatti and Lassad Elmoubarki and Abdeltawab Hendawi},
      year={2025},
      eprint={2505.09598},
      archivePrefix={arXiv},
      primaryClass={cs.CY},
      url={https://arxiv.org/abs/2505.09598}, 
}

@article{wang2021combining,
author = {Wang, Hongwei and Leskovec, Jure},
title = {Combining Graph Convolutional Neural Networks and Label Propagation},
year = {2021},
issue_date = {October 2022},
publisher = {Association for Computing Machinery},
address = {New York, NY, USA},
volume = {40},
number = {4},
issn = {1046-8188},
url = {https://doi.org/10.1145/3490478},
doi = {10.1145/3490478},
month = nov,
journal = {ACM Trans. Inf. Syst.},
articleno = {73},
numpages = {27}
}

@misc{grattafiori2024thellama3,
      title={The Llama 3 Herd of Models}, 
      author={Aaron Grattafiori and Abhimanyu Dubey and Abhinav Jauhri and Abhinav Pandey and Abhishek Kadian and Ahmad Al-Dahle and Aiesha Letman and Akhil Mathur and Alan Schelten and Alex Vaughan and Amy Yang and Angela Fan and others},
      year={2024},
      eprint={2407.21783},
      archivePrefix={arXiv},
      primaryClass={cs.AI},
      url={https://arxiv.org/abs/2407.21783}, 
}

@misc{hu2024letsaskgnn,
  title = {Let's {{Ask GNN}}: {{Empowering Large Language Model}} for {{Graph In-Context Learning}}},
  shorttitle = {Let's {{Ask GNN}}},
  author = {Hu, Zhengyu and Li, Yichuan and Chen, Zhengyu and Wang, Jingang and Liu, Han and Lee, Kyumin and Ding, Kaize},
  year = {2024},
  publisher = {arXiv},
  doi = {10.48550/ARXIV.2410.07074},
  urldate = {2025-08-28},
  copyright = {arXiv.org perpetual, non-exclusive license}
}

@inproceedings{ye2024language,
    title = "Language is All a Graph Needs",
    author = "Ye, Ruosong  and
      Zhang, Caiqi  and
      Wang, Runhui  and
      Xu, Shuyuan  and
      Zhang, Yongfeng",
    editor = "Graham, Yvette  and
      Purver, Matthew",
    booktitle = "Findings of the Association for Computational Linguistics: EACL 2024",
    month = mar,
    year = "2024",
    address = "St. Julian{'}s, Malta",
    publisher = "Association for Computational Linguistics",
    url = "https://aclanthology.org/2024.findings-eacl.132/",
    pages = "1955--1973",
}

\end{document}